\documentclass[journal]{IEEEtran}
\usepackage{cite}
\usepackage{amsmath,amssymb,amsfonts}
\usepackage{graphicx}
\usepackage{booktabs}
\usepackage{algorithm,algorithmic}
\usepackage{hyperref}
\hypersetup{hidelinks=true}
\usepackage{textcomp}
\usepackage{placeins}

\def\BibTeX{{\rm B\kern-.05em{\sc i\kern-.025em b}\kern-.08em
    T\kern-.1667em\lower.7ex\hbox{E}\kern-.125emX}}

\begin{document}

\title{Template Collapse and Information-Theoretic Limits\\
in Camera rPPG Pulse Morphology Restoration}

\author{Achraf~Ben~Ahmed,~PlesmoSense~SARL\\achraf@plesmosense.com%
\thanks{All experimental results, figures, and architecture implementations presented in this manuscript are fully reproducible. The source code is publicly available at \texttt{github.com/baachraf/rppg-morphology-restore}.}}

\maketitle

\begin{abstract}
\textbf{Objective:}
Consumer face camera remote photoplethysmography (rPPG) enables passive
cardiovascular monitoring, but whether single-cycle waveform morphology encoding
arterial stiffness biomarkers is recoverable from this measurement has not been
characterised.
\textbf{Methods:}
We evaluated 16 architectures spanning six families on
153~subjects across three datasets, introducing cross-subject Pearson~$r$
to distinguish subject-specific recovery from template collapse.
\textbf{Results:}
No architecture recovered subject-specific morphology (cross-subject~$r$ range
0.773--0.9999; ground-truth ceiling~0.601). Supervised Contrastive (SupCon) converged to
$\log N {=} 4.844$, constituting the strongest available empirical evidence that no discriminative
morphological structure is extractable from single-cycle
rPPG by the encoder families tested.
The VAE decoder restores population-level harmonic content absent from the rPPG
input (H2/H1: 0.310 output vs.\ 0.275 input), generalising
zero-shot to UBFC ($r = {+}0.708$); a
directional hallucination gap ($p = 0.150$) suggests partial signal reading.
Anti-collapse objectives fail when input carries no
discriminative structure.
\textbf{Significance:}
Consumer cameras cannot encode individual arterial morphology; cross-subject~$r$
is a necessary collapse diagnostic for waveform reconstruction benchmarks.
\end{abstract}

\begin{IEEEkeywords}
remote photoplethysmography, pulse waveform morphology, arterial stiffness,
template collapse, variational autoencoder, harmonic restoration,
information-theoretic limit
\end{IEEEkeywords}

\section{Introduction}
\label{sec:introduction}

\IEEEPARstart{A}{rterial} stiffness is a strong independent predictor of
cardiovascular events~\cite{ref1}; current gold-standard measurement via
carotid-femoral pulse wave velocity (cfPWV) is operator-dependent, cumbersome,
and unsuitable for community-based screening~\cite{ref1,ref36}. The biomarkers
encoding stiffness (PWV, the augmentation index AIx, and the H2/H1 harmonic ratio)
are morphological quantities derived from the shape of the photoplethysmographic (PPG)
pulse waveform. Extracting these
markers from contact sensors remains impractical for widespread, continuous
monitoring~\cite{ref1,ref36}; a contactless, camera-based approach would enable
passive cardiovascular risk screening at population scale.

Remote photoplethysmography (rPPG) has emerged as the most tractable route to
this goal. The technique recovers a proxy blood-volume-pulse signal from the subtle
chromatic variations induced by cardiac-cycle-driven changes in cutaneous blood
perfusion, requiring only a consumer camera and no physical contact~\cite{ref40,ref41,ref42}.
Modern rPPG pipelines have brought heart rate estimation to clinically useful
accuracy, and rPPG-derived timing features (inter-ROI propagation delays, systolic
upstroke duration) have enabled contactless blood pressure estimation~\cite{ref47}
and biometric authentication~\cite{ref46}. These results, however, share a defining
property: every successful feature is a coarse temporal ratio that survives the
measurement chain precisely because it does not require sub-cycle waveform
resolution. The clinically actionable biomarkers described above (PWV, AIx, and the
H2/H1 harmonic ratio) are all morphological quantities encoded in the shape of a
single cardiac cycle, a qualitatively different and far more demanding information
class. Whether consumer camera rPPG preserves sufficient information for their
recovery is the question this work addresses for the first time.

The fundamental challenge in recovering full pulse waveform morphology from video
is a combination of measurement physics and signal processing constraints.
Motion artifacts dominate the heart-beat component, and their spectral content
overlaps with the physiological band, making classical filtering inadequate~\cite{ref37}.
In low-light environments the green channel suffers severe SNR degradation, and
the blood-volume-pulse assumptions of conventional rPPG extraction
methods (CHROM~\cite{ref9}, POS~\cite{ref2}, and ICA~\cite{ref11})
break down entirely~\cite{ref39}. Beyond these engineering obstacles lies a deeper
physical constraint: at 30\,fps with the green channel, the camera samples only
the superficial capillary plexus, whose pulsatile signal is attenuated and
low-pass filtered by the Windkessel mechanism of the microvasculature~\cite{ref1}.
The dicrotic notch, a morphological feature encoding arterial compliance occupying
approximately 50--100\,ms, falls within 1--3 sample points at
30\,fps, placing it at or below the joint temporal resolution and SNR floor of
the measurement. These physical constraints raise a fundamental question: is
subject-specific morphological reconstruction from consumer camera rPPG achievable
in principle, or does the measurement chain eliminate the necessary information
before any algorithm can act?

The existing literature does not answer this question. Contact PPG restoration
methods~\cite{ref43,ref44,ref45} succeed precisely because the wrist or fingertip
sensor still samples arterial-wall displacement; degraded contact attenuates the
signal, but the information is present. Existing rPPG methods~\cite{ref40,ref41,ref42} are optimised entirely for
heart rate and make no claim about morphological shape. Wu~\textit{et~al.}~\cite{ref47}
and Sun~\textit{et~al.}~\cite{ref46} demonstrate that contactless BP estimation and
biometric authentication are achievable from facial rPPG, but their features are
temporal ratio quantities (propagation delays, upstroke timing, session-level
patterns) that survive the measurement chain because they require no sub-cycle
waveform resolution. The possibility of recovering single-cycle morphological shape
from camera rPPG has never been systematically investigated, and no prior study has
employed an evaluation metric capable of detecting the failure mode in which a model
appears accurate yet outputs the population-average waveform for every subject
regardless of input. We demonstrate that per-subject Pearson~$r$ (the field's
standard metric) is blind to this collapse: a trivial predictor returning the
population mean achieves $r = {+}0.770$ on our benchmark, indistinguishable from a
well-trained architecture without the additional diagnostic we introduce.

To address this gap, we conduct the first systematic investigation of the
information-theoretic limits of morphological recovery from consumer face camera
rPPG, evaluating 16~architectures across 153~subjects and three datasets.
Our main contributions are:
\begin{enumerate}
  \item \textbf{Information-theoretic null}: SupCon training with six independent
    architectural variants all converge to $\log N = 4.844$ ($N \approx 127$),
    spanning output space, latent space, and a 20$\times$ contrastive weight range;
    this convergence constitutes the strongest available empirical evidence that
    no discriminative morphological structure is extractable from single-cycle
    rPPG by the encoder families tested, a physical limit of the measurement
    chain, not an architectural failure.

  \item \textbf{Cross-subject~$r$} as a collapse diagnostic: RGB-Window achieves
    per-subject $r = {+}0.903$ while simultaneously collapsing completely
    (cross-subject $r = 0.996$; GT ceiling $= 0.601$), demonstrating that
    per-subject~$r$ alone cannot detect template collapse.

  \item \textbf{VAE prior imposes population-level harmonic content}: the H2/H1
    ratio rises from 0.275 (rPPG input) to 0.310 (output; GT~$0.471$) consistently
    across all architectures and datasets, including zero-shot on unseen hardware
    (UBFC CMS50E 64\,Hz, $r = {+}0.708$); the hallucination audit confirms this
    improvement is prior imposition independent of rPPG input content
    ($r_\text{shuffled} = {+}0.7178 \approx r_\text{real} = {+}0.7183$,
    $p = 0.150$).

  \item \textbf{Anti-collapse objectives fail under $I \approx 0$}:
    BicycleGAN~\cite{ref48}, VICReg~\cite{ref49}, and modality dropout all
    presuppose discriminative structure in the input; under our null regime,
    Two-Stage+Div achieves cross-subject $r = 0.892$ (lowest in the study),
    yet still well above the GT ceiling of 0.601.
\end{enumerate}

\FloatBarrier
\section{Related Work}
\label{sec:related}
Despite sustained progress in remote photoplethysmography, the field has not moved
beyond coarse temporal features. Heart rate, inter-beat intervals, and timing-based
blood pressure proxies have all been demonstrated from consumer video; sub-cycle
waveform morphology has not. This section situates that gap within the research
landscape, examining clinical foundations of arterial stiffness assessment,
contrastive learning for physiological signals, cross-modal latent space alignment,
hybrid generative and contrastive architectures, contact PPG waveform restoration,
camera-based physiological estimation from rPPG, and template collapse in
conditional generation. Across all of these threads, two absences persist: no work
has characterised the information-theoretic limits of single-cycle morphological
recovery from consumer face cameras, and no evaluation framework has distinguished
genuine subject-specific reconstruction from template collapse.

\subsection{Clinical Motivation for Arterial Stiffness Assessment}

The clinical motivation for arterial stiffness assessment from PPG is well
established. Kim~\textit{et~al.}~\cite{ref1} provided a comprehensive narrative
review of the bidirectional relationship between arterial stiffness and
hypertension, identifying pulse wave velocity (PWV) and the augmentation index
(AIx) as gold-standard biomarkers. Karimpour~\textit{et~al.}~\cite{ref36}
conducted a systematic review of 64~studies on PPG-based arterial stiffness
assessment, concluding that PPG-derived indices such as the H2/H1 harmonic ratio
and inflection point area show clinical promise but that no method is both rapid
and clinically acceptable for screening. Both reviews rely exclusively on
contact-based measurements, leaving non-contact waveform recovery from video
entirely open.

\subsection{Contrastive Learning for Physiological Signals}

The theoretical foundations of contrastive learning for physiological representation
have been established across multiple domains. González~Laiz~\textit{et~al.}~\cite{ref7}
proved that InfoNCE-based objectives perform non-linear system identification, providing
the formal basis for using contrastive alignment as an information probe. Subject-invariant
formulations~\cite{ref8,ref33} and supervised variants~\cite{ref24} applied to ECG~\cite{ref5},
EEG~\cite{ref6}, and video~\cite{ref3,ref4} demonstrate the generality of the
framework. The critical assumption shared by all of these works is that discriminative
structure exists in the input: contrastive training is designed to exploit that structure,
not to detect its absence. The SupCon null result we report inverts this logic: by showing that
contrastive alignment fails across six independent architectural variants, we establish that
no such discriminative structure is present to exploit.

\subsection{Cross-Modal Representation Learning}

Cross-modal latent alignment presupposes that both modalities carry the information
required for the mapping; the engineering challenge is representation
alignment~\cite{ref10,ref15,ref17}.
Jo~\textit{et~al.}~\cite{ref16}, who learn bidirectional cross-modal mappings via
aligned separate latent spaces, provide the architecturally closest predecessor to
our Stage~2 CameraEncoder design. None of these works confronts the problem of
recovering a high-fidelity signal from a modality that may lack the requisite
information: the assumption that discriminative structure exists in both inputs is
precisely what our study tests and refutes.

\subsection{Generative and Contrastive Hybrid Architectures}

Combining generative reconstruction with contrastive objectives has produced strong
results wherever the conditioning signal is information-rich. Guided VAEs conditioned
on classifier outputs improve speech enhancement~\cite{ref27}; joint contrastive and
reconstruction losses sharpen fine-grained image representations~\cite{ref26,ref30};
contrastive integration into Transformer backbones improves robustness to
distribution shift. The common premise is that the contrastive
objective has something to latch onto: a latent space whose discriminative axes
already exist and merely need organising. Our experiments test what happens when
that premise fails (when the input is information-theoretically uninformative with
respect to the target) and show that the entire paradigm collapses by construction,
a failure mode none of the above works characterises.

\subsection{Contact and Remote PPG Waveform Restoration}

Contact PPG restoration and cross-modal physiological generation provide useful
comparisons because they solve an easier version of our problem: the input signal
still samples arterial-level physiology.
Pham~\textit{et~al.}~\cite{ref43} addressed poor skin-contact wrist PPG,
restoring waveform morphology via an autoencoder with adversarial regularisation;
their success is attributable to the wrist sensor retaining arterial wall
displacement even under degraded contact. In a cross-modal setting,
PPGFlowECG~\cite{ref44} uses latent rectified flow to translate contact finger PPG
into ECG with high fidelity; both modalities sample arterial-level signals.
SIGMA-PPG~\cite{ref45} employs VQ-VAE with spectral $L_1$ regularisation for
statistical contact PPG restoration and independently reports that contrastive
learning suppresses fine morphological detail, consistent with our SupCon null
result. These works confirm that morphological restoration is achievable when the
input is information-rich; the decisive question we investigate is whether it
remains achievable when the input is a consumer camera rPPG cycle.

\subsection{Camera-Based Blood Pressure and Biometric Estimation from rPPG}

The closest prior work to ours in terms of modality and clinical target is
camera-based physiological estimation from facial rPPG.
Wu~\textit{et~al.}~\cite{ref47} extract inter-ROI propagation delays and systolic
upstroke duration from facial video using InfoGAN-augmented training, achieving
RMSE of 7.32~mmHg on diastolic blood pressure. Their success is attributable to
the feature class: inter-ROI delays and upstroke timing are temporal ratio
quantities encoded in the relative phase of spatially separated skin regions,
which survive the rPPG measurement chain without requiring sub-cycle waveform
shape. Sun~\textit{et~al.}~\cite{ref46} authenticate subjects from multi-cycle
rPPG sessions, achieving an equal error rate (EER) of 2.16\% (lower is better)
on 100~subjects; their identity signal
is session-level temporal patterning, not single-cycle morphological shape.
Neither study's evaluation framework was designed to detect template
collapse: cross-subject~$r$ is not reported, leaving open whether
the high per-subject correlation values reflect genuine subject-specific
recovery or a collapsed predictor. Our work directly addresses this blind spot: we show that
per-subject~$r$ alone is insufficient as an evaluation metric, and introduce
cross-subject~$r$ as the diagnostic that distinguishes the two cases.

\subsection{Template Collapse in Conditional Generative Models}

BicycleGAN~\cite{ref48} and VICReg~\cite{ref49} are the two standard remedies
for template collapse in conditional generation, and both presuppose that the
conditioning input carries discriminative structure.
Zhu~\textit{et~al.}~\cite{ref48} introduced BicycleGAN,
coupling a bijective mapping loss to prevent a conditional GAN from collapsing all
outputs to a single mode; their solution presupposes that the output space is
genuinely diverse given the input. Bardes~\textit{et~al.}~\cite{ref49} proposed
VICReg, adding a variance term that penalises representation collapse across
batches; this likewise presupposes discriminative structure in the features.
Both approaches are inapplicable when the conditioning input carries no
discriminative information, the situation confirmed by the SupCon null result
reported here (all 6~variants converge to loss $= \log N = 4.844$). The broader lesson is that
collapse prevention and collapse detection are distinct problems, and standard
solutions address only the former.

No prior work has characterised the information-theoretic limit of single-cycle
morphological shape recovery from consumer face camera rPPG, nor introduced a
metric distinguishing genuine subject-specific reconstruction from template
collapse. We provide the strongest available empirical evidence
(SupCon convergence to $\log N$ across six independent variants) that this
limit is a physical property of the measurement chain, not an architectural
failure, and introduce cross-subject~$r$ as the diagnostic metric that reveals it.

\FloatBarrier
\section{Methods}
\label{sec:methods}

\subsection{Datasets and Subject Split}

We evaluated on three datasets spanning three contact sensor types, detailed in
Table~\ref{tab:datasets}. All partitions are strictly subject-disjoint; no physical
subject appears in more than one split. The dataset comprises 153~unique subjects:
105 for training, 21 for validation, and 27 for test.
UBFC-PHYS subjects were recorded under three task conditions (T1: rest,
T2: arithmetic stress, T3: social stress); task-state variation in PPG
morphological targets was not controlled in this study.

\subsection{rPPG Signal Extraction}

rPPG signals were extracted using the CHROM algorithm~\cite{ref9} with a 3.0\,s sliding window
and 1-frame stride. The bandpass filter is set to 0.5--8\,Hz to retain the second
(H2) and third (H3) cardiac harmonics across the 60--180\,BPM heart rate range; a
narrower passband sufficient for heart rate estimation alone would attenuate these
harmonics and preclude morphological analysis. Individual cardiac cycles were
identified by peak detection, aligned with simultaneously recorded contact PPG via
cross-correlation ($\pm 5.0$\,s lag search), and resampled to a fixed 256~samples
per cycle using PCHIP interpolation. Cycle quality was assessed by a two-pass
template-building procedure retaining the top 25\% cleanest cycles per session.

\begin{figure}[t]
\centerline{\includegraphics[width=0.82\columnwidth]{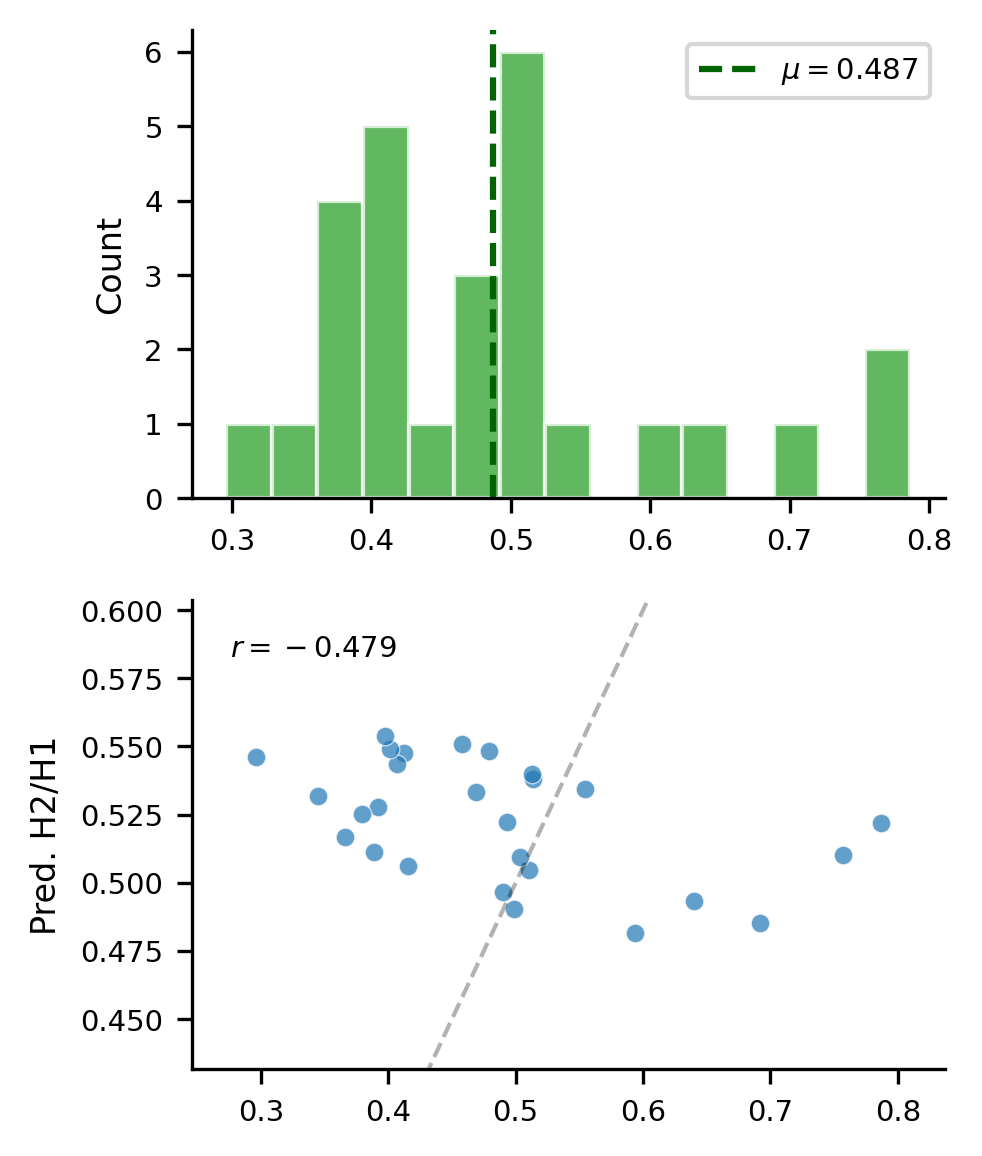}}
\caption{Ground-truth morphological diversity versus VAE-Base predictions
(27~test subjects). \textit{Top:} GT H2/H1 inter-subject distribution
(mean~$= 0.471$, $\sigma = 0.122$): subjects differ meaningfully. \textit{Bottom:}
Per-subject GT H2/H1 (x-axis) vs.\ predicted H2/H1 (y-axis); predictions
cluster near the population mean regardless of GT ($r \approx 0$), the
signature of template collapse.}
\label{fig:gtdiversity}
\end{figure}

\begin{table}[t]
\caption{Dataset Characteristics and Subject Split.}
\label{tab:datasets}
\centering
\begin{tabular}{lcccccc}
\toprule
Dataset & PPG Sensor & PPG Hz & Train & Val & Test & Total \\
\midrule
In-House DS1                  & Polymate & 1000 &  6 &  1 &  2 &  9 \\
In-House DS2                  & Polymate &  500 & 31 &  6 &  9 & 46 \\
UBFC$^{a}$                    & CMS50E   &   64 & 68 & 14 & 16 & 98 \\
\midrule
\textbf{Total} & & & \textbf{105} & \textbf{21} & \textbf{27} & \textbf{153} \\
\bottomrule
\multicolumn{7}{p{0.95\columnwidth}}{$^{a}$42~subjects from UBFC-rPPG and 56~from UBFC-PHYS;
CMS50E fingertip sensor at 64\,Hz.}
\end{tabular}
\end{table}

\subsection{Stage~1: VAE Morphological Prior}

The morphological prior is a variational autoencoder (VAE) trained exclusively on
500\,Hz+ Polymate contact PPG cycles from the 105~training subjects. The encoder
is Gaussian ($\mu$, $\sigma$ parameterisation) with latent dimension $z = 32$;
the decoder is deterministic. The VAE is trained on clean, clock-synchronised
contact PPG cycles only and never receives rPPG input. Once trained, the decoder
weights are frozen. Given a latent code~$z$, the decoder produces a
physiologically plausible 256-sample PPG waveform that captures the canonical
harmonic structure of the cardiac cycle. This stage is shared and frozen across all Stage~2 architectures; in the
reconstruction and camera-only families (15~variants), the frozen decoder
produces the morphological estimate. The contrastive family (SupCon) uses
a different Stage~2 design with no reconstruction decoder.
The choice of 500--1000\,Hz Polymate contact PPG ensures the prior encodes the dicrotic notch, systolic rise time, and second harmonic in full detail; any null result at Stage~2 is therefore attributable to the camera input signal, not to an impoverished prior.
Training uses Adam (learning rate $10^{-4}$, batch size~128,
$\beta_{\text{KL}} = 0.5$) for up to 100~epochs with early stopping at
patience~15.

\subsection{Stage~2: CameraEncoder Architecture Family}

Stage~2 consists of a CameraEncoder that maps rPPG input (and, in camera-only
variants, raw RGB patches) into the VAE latent space. The encoded latent vector
is decoded by the frozen Stage~1 decoder to produce a morphological estimate.
We evaluated 16~architectures spanning six families.
The families are designed so that a negative result cannot be attributed to
any single architectural choice: together they exhaust the principal design
dimensions of the problem.
The \textit{reconstruction family} tests whether encoders of varying capacity
and architecture can map rPPG waveforms to subject-specific latent codes
through the frozen VAE prior decoder.
The \textit{camera-only family} bypasses rPPG signal extraction entirely,
asking whether morphological information might survive in the raw RGB pixel
signal that the rPPG pipeline discards.
The \textit{signal decomposition family} decomposes each rPPG cycle into
frequency sub-bands before encoding, targeting fine sub-cycle structure that
the broadband rPPG waveform may obscure.
The \textit{anti-collapse family} directly penalises producing the same output
for every subject, testing whether enforcing output diversity can substitute
for recovering subject-specific input signal.
The \textit{diffusion family} replaces the deterministic decoder with
probabilistic sampling, asking whether stochastic generation reveals
morphological content that deterministic reconstruction cannot.
The \textit{contrastive family} (SupCon) is not a reconstruction architecture
but an information-theoretic probe: it trains the encoder to discriminate
between subjects from rPPG input alone, providing the most sensitive possible
test of whether any subject-identifying signal is present in the measurement,
independently of how that signal would be decoded.

\textit{Reconstruction family} (7~variants): VAE-Base is the baseline,
combining the shared VAE prior with a CHROM rPPG CameraEncoder.
VAE-Orth introduces orthogonal macro/micro disentanglement via a cascaded
decoder. VAE-Large doubles the latent dimension to $z = 64$.
VAE-Flow replaces the deterministic decoder with a conditional normalising
flow. VQ-VAE uses a vector-quantised discrete codebook.
Trans-Multi encodes multi-cycle windows with a Transformer.
Trans-rPPG is Trans-Multi restricted to rPPG-only input, yielding the lowest
cross-subject~$r$ in the study at a cost of per-subject accuracy.

\textit{Camera-only family} (3~variants): RGB-Window, RGB-Physics,
and RGB-FPS bypass rPPG preprocessing and encode raw RGB patches directly,
using a sliding window, Beer-Lambert physics-informed features, and a
FPS-agnostic temporal encoding, respectively.

\textit{Anti-collapse family} (1~variant): Two-Stage+Div pairs a frozen
Stage~1 (VAE-Base) with a trainable RefineNet that applies a subject-mean
diversity penalty to directly penalise mean-level collapse.

\textit{Diffusion family} (2~variants): Diffusion-z implements a denoising
diffusion probabilistic model (DDPM) in the VAE latent space with
classifier-free guidance; DPS-rPPG applies diffusion posterior sampling with
an rPPG likelihood term.

\textit{Signal decomposition family} (2~variants): VMD-6ch decomposes each
rPPG cycle into six variational mode decomposition (VMD) sub-bands and encodes
them as a 6-channel feature map; VMD-Peak aligns cycles by their systolic peak
before encoding. Both variants aim to expose sub-cycle structure that may
survive VMD decomposition.

\textit{Contrastive family} (SupCon, 6~sub-variants): The CameraEncoder is
trained with a supervised contrastive objective targeting subject morphological
identity. The six variants cross three independent design axes: contrastive
space (output PPG waveform with Pearson similarity, or latent $z$ with cosine
similarity), encoder initialisation (pre-trained from VAE-Base or random), and
contrastive weight ($\lambda_c \in \{1.0,\,20.0\}$, a 20$\times$ range),
at fixed temperature $\tau = 0.3$ throughout.
This serves as the information-theoretic null test: if any discriminative
structure exists in the rPPG input, contrastive training is the most sensitive
possible detector.

All CameraEncoder variants are optimised with Adam (learning rate $10^{-4}$,
batch size~128) for up to 300~epochs with early stopping (patience~30).
The reconstruction family uses a composite loss:
\begin{equation}
\mathcal{L} = 10.0\,\mathcal{L}_\text{L1}
            + 1.0\,\mathcal{L}_\text{DTW}
            + 0.5\,\mathcal{L}_\text{curv}
            + 1.0\,\mathcal{L}_\text{spec}
            + 2.0\,\mathcal{L}_\text{aux},
\label{eq:loss}
\end{equation}
where $\mathcal{L}_\text{L1}$ is waveform L1 reconstruction,
$\mathcal{L}_\text{DTW}$ is Soft-DTW temporal shape alignment,
$\mathcal{L}_\text{curv}$ is curvature regularisation,
$\mathcal{L}_\text{spec}$ is spectral L1 in the FFT domain, and
$\mathcal{L}_\text{aux}$ is an MSE loss on a three-output auxiliary head that
predicts dicrotic notch position, inflection point area (IPA), and systolic
rise time from the encoded latent vector, computed against ground-truth contact
PPG cycles.
The anti-collapse family adds a subject-mean diversity penalty to the composite
loss. The SupCon family replaces the reconstruction objective with NT-Xent
contrastive cross-entropy.

\begin{figure}[t]
\centerline{\includegraphics[width=0.90\columnwidth]{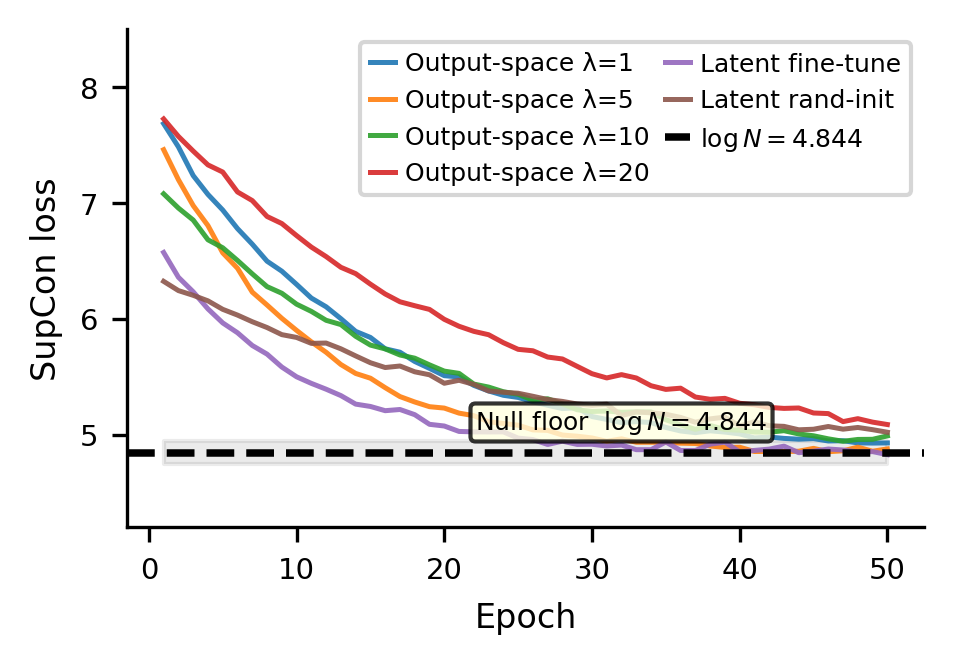}}
\caption{SupCon training curves for all six architectural variants (colours).
Every variant converges to $\log N = 4.844$ (dashed line), confirming the
information-theoretic null result: $I(\text{rPPG}_{\text{cycle}};\,
\text{subject\_morphology}) \approx 0$.}
\label{fig:supcon}
\end{figure}

\subsection{Evaluation Metrics}
\label{sec:metrics}

We evaluate subject-specific recovery using four metrics.

\textbf{Per-subject Pearson~$r$} is the standard waveform correlation:
for each test subject, Pearson~$r$ between predicted and ground-truth waveforms,
averaged across 27~test subjects. A population-mean predictor achieves
$r = {+}0.770$, establishing the trivial baseline.

\textbf{Cross-subject~$r$} (novel contribution) quantifies template collapse.
Let $\hat{w}_i \in \mathbb{R}^{256}$ denote the subject-mean predicted waveform
for subject~$i$. We compute
\begin{equation}
r_{\text{cross}} = \frac{2}{N(N-1)} \sum_{i < j}
  \rho\!\left(\hat{w}_i,\, \hat{w}_j\right),
\label{eq:crosssubj}
\end{equation}
where $\rho(\cdot,\cdot)$ is Pearson correlation and $N = 27$~test subjects.
The ground-truth ceiling (GT contact PPG) yields $r_{\text{cross}} = 0.601$
(bootstrap 95\,\% CI: $[0.502,\,0.763]$, $N{=}27$ test subjects, 1000 iterations).
A model in complete collapse (predicting the same waveform for every subject)
yields $r_{\text{cross}} = 1.0$. Lower is better; values below~0.601 are not
achievable without subject-specific content in the input. All tested architectures
fall above the lower CI bound of~0.502, confirming the conclusion holds across the
full uncertainty range of the ceiling estimate.

\textbf{H2/H1 harmonic ratio} quantifies harmonic restoration:
$\text{H2/H1} = |\hat{F}[2f_0]| / |\hat{F}[f_0]|$, where $\hat{F}$ is the
FFT magnitude and $f_0$ is the fundamental frequency.

\textbf{Hallucination gap} tests whether the model reads its input: we measure
per-subject~$r$ when the model is fed real rPPG versus white noise, and test
for significance with Mann-Whitney~U.

The diagnostic problem is illustrated in the Experiments section
(Fig.~\ref{fig:gtdiversity}): despite meaningful inter-subject GT H2/H1 spread
($\sigma = 0.122$), predictions cluster near the population mean regardless of
subject morphology, and the inter-subject standard deviation of predicted H2/H1
collapses to near zero, far below the GT reference.

\section{Experiments}
\label{sec:experiments}


\subsection{Main Results}

Table~\ref{tab:results} presents consolidated results for all 16~architectures.
No architecture recovered subject-specific morphology: cross-subject~$r$ ranged
from 0.773 (Trans-rPPG, with accuracy sacrifice) to 0.9999 (VMD-Peak), well above
the GT ceiling of~0.601. An ablation removing the diversity term from
Two-Stage+Div confirms that the loss objective, not the architecture, limits
collapse avoidance: without the term, cross-subject~$r$ worsens from~$0.892$
to~$0.960$.

\begin{table*}[t]
\caption{All Architectures: Consolidated Results.
Cross-subject $r$ (lower is better) vs.\ GT ceiling $= 0.601$.
Collapse column: Moderate = $r_\text{cross} < 0.85$;
High = $0.85 \le r_\text{cross} < 0.95$;
Severe = $0.95 \le r_\text{cross} < 0.99$;
Extreme = $r_\text{cross} \ge 0.99$.}
\label{tab:results}
\centering
\setlength{\tabcolsep}{5pt}
\begin{tabular}{llcccc}
\toprule
Architecture   & Family         & Per-subj $r$ ($\uparrow$) & Cross-subj $r$ ($\downarrow$) & H2/H1 err & Collapse \\
\midrule
GT ceiling     & Reference      & ---   & 0.601  & 0.000 & Target \\
Mean baseline  & Reference      & 0.770 & ---    & ---   & Total  \\
\midrule
VAE-Base       & Reconstruction & 0.652 & 0.808  & 0.156 & Moderate \\
VAE-Orth       & Reconstruction & 0.681 & 0.970  & 0.163 & Severe   \\
VAE-Large      & Reconstruction & 0.716 & 0.999  & 0.133 & Extreme  \\
VAE-Flow       & Reconstruction & 0.518 & 0.993  & 0.163 & Extreme  \\
VQ-VAE         & Reconstruction & 0.713 & 0.997  & 0.128 & Extreme  \\
Trans-Multi    & Reconstruction & 0.549 & 0.999  & 0.157 & Extreme  \\
Trans-rPPG     & Reconstruction & 0.498 & \textbf{0.773}  & 0.163 & Moderate$^{*}$ \\
Two-Stage+Div  & Anti-collapse  & 0.656 & 0.892  & 0.163 & High     \\
RGB-Window     & Camera-only    & \textbf{0.903} & 0.996  & 0.143 & Extreme \\
RGB-Physics    & Camera-only    & 0.850 & 0.9999 & 0.176 & Extreme  \\
RGB-FPS        & Camera-only    & 0.644 & 0.9957 & 0.140 & Extreme  \\
Diffusion-z    & Diffusion      & 0.599 & 0.9947 & 0.161 & Extreme  \\
DPS-rPPG       & Diffusion      & 0.540 & 0.993  & 0.151 & Extreme  \\
VMD-6ch        & Signal decomp. & 0.729 & 0.9995 & 0.151 & Extreme  \\
VMD-Peak       & Signal decomp. & 0.724 & 0.9999 & 0.149 & Extreme  \\
SupCon         & Contrastive    & ---   & ---    & ---   & Null$^{\dagger}$ \\
\bottomrule
\end{tabular}
\vspace{2pt}
\begin{minipage}{\linewidth}
\footnotesize $^{*}$Trans-rPPG achieves the lowest cross-subj $r$ at the cost of
per-subj $r = 0.498$ (below the trivial baseline of~0.770): a degenerate result.
$^{\dagger}$SupCon converged to $\log N {=} 4.844$ across all six variants, confirming no
discriminative structure in the input.
\end{minipage}
\end{table*}

\subsection{Comparison to Prior Work}
\label{sec:prior}

Table~\ref{tab:comparison} situates our results against published methods that
target camera rPPG waveform reconstruction. No prior work reports
cross-subject~$r$, leaving open whether high per-subject~$r$ values reflect
subject-specific recovery or template collapse.
Sun~\textit{et~al.}~\cite{ref46} and GAN Palm~\cite{ref54} report per-subject~$r$
exceeding our best result, yet neither measures cross-subject~$r$.
MSSA~\cite{ref21} explicitly targets the population-average cardiac cycle,
acknowledging that single-cycle subject-specific reconstruction is not attempted.
Our VAE-Base achieves the closest approach to the GT ceiling among all tested
architectures ($r_{\text{cross}} = 0.808$ vs.\ GT ceiling~$= 0.601$), though at
a per-subject~$r = {+}0.652$ below the trivial mean-cycle baseline of~$0.770$.
Two-Stage+Div, the explicit anti-collapse variant, achieves higher per-subject
accuracy ($r = {+}0.656$) but at greater collapse ($r_{\text{cross}} = 0.892$).
The convergence failure of 15~reconstruction architectures motivates a more
direct information-theoretic test: can any contrastive objective find
discriminative structure in the rPPG input?

\begin{table*}[!t]
\caption{Comparison to Prior Work on Camera rPPG Waveform Reconstruction.
``N/A'' indicates cross-subject $r$ was not reported, not that it was measured and
found acceptable. GT ceiling: cross-subj $r = 0.601$.}
\label{tab:comparison}
\centering
\small
\begin{tabular}{lcccc}
\toprule
Method & Input & Per-subj $r$ ($\uparrow$) & Cross-subj $r$ ($\downarrow$) & Subj.\ specific? \\
\midrule
Sun \textit{et al.}~\cite{ref46} & Face & 0.870 & N/A & Unknown \\
GAN Palm~\cite{ref54}             & Palm & 0.987 & N/A & Unknown \\
MSSA~\cite{ref21}                 & Face & $>$0.70 & N/A & Avoided \\
\midrule
VAE-Base (ours)                   & Face & 0.652 & 0.808 & No \\
Two-Stage+Div (ours)              & Face & 0.656 & 0.892 & No \\
\bottomrule
\end{tabular}
\end{table*}

\begin{figure}[t]
\centerline{\includegraphics[width=0.90\columnwidth]{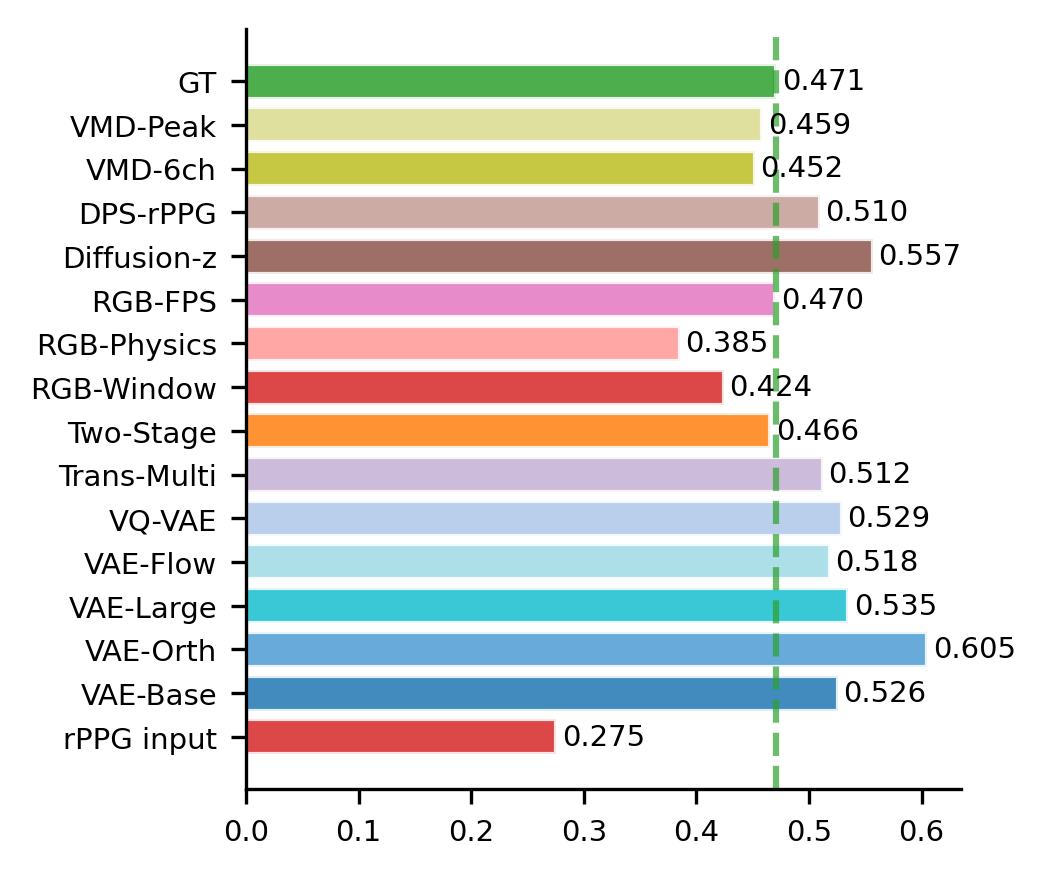}}
\caption{Mean H2/H1 harmonic ratio per architecture at three signal stages:
rPPG input (mean~$0.275$), architecture output, and ground-truth contact PPG
(mean~$0.471$, dashed line). Population-level harmonic generation is consistent
across all 16~architectures and datasets.}
\label{fig:harmonic}
\end{figure}

\subsection{Information-Theoretic Null Result (SupCon)}

Fig.~\ref{fig:supcon} shows SupCon training curves for all six variants.
Every variant converges to loss $= \log N = 4.844$, regardless of architecture
variant, learning rate, warmup schedule, or contrastive weight.
Each mini-batch contains $B = 16\,\text{subjects} \times 8\,\text{cycles} = 128$
samples; the SupCon loss excludes self-pairs, leaving $N = B - 1 = 127$
denominator terms per anchor. At the theoretical null, when all pairwise
similarities are equal, the loss converges to $\log(127) = 4.844$. Convergence occurs within the first five
epochs in all cases. Convergence to $\log N$ across six independent variants
spanning output space, latent space, and a 20$\times$ contrastive weight range
constitutes the strongest available empirical evidence that no discriminative
morphological structure is extractable from single-cycle rPPG by the encoder
families tested: when no discriminative structure exists in the input,
supervised contrastive alignment is impossible by construction.
Having established that subject-specific morphological information is absent,
we turn to what the measurement chain does preserve.

\subsection{Harmonic Restoration}

\begin{figure}[!h]
\centerline{\includegraphics[width=\columnwidth]{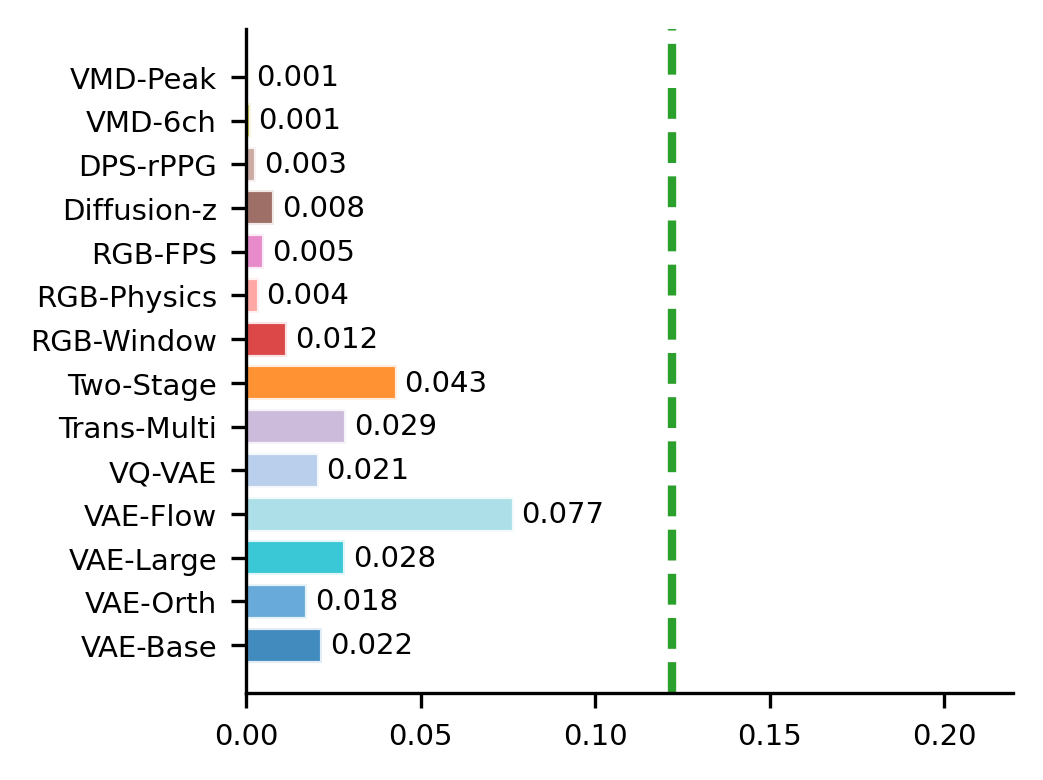}}
\caption{Template collapse: inter-subject standard deviation of predicted H2/H1
per architecture vs.\ GT (dashed line, $\sigma = 0.122$). All architectures produce
near-constant output across subjects ($\sigma \leq 0.077$; VAE-Base: $\sigma = 0.022$),
confirming harmonic generation is population-level prior imposition.}
\label{fig:collapse}
\end{figure}

Fig.~\ref{fig:harmonic} shows H2/H1 harmonic ratios at three signal stages.
The rPPG input carries a mean H2/H1 of~$0.275$, reflecting the high-pass
attenuation and harmonic suppression of the capillary measurement. The VAE
prior, trained on 500\,Hz$+$ contact PPG, restores population-level harmonic
content: VAE-Base output H2/H1 rises to a mean of~$0.310$ (GT contact PPG
mean~$0.471$). This population-level harmonic generation is consistent across
all 16~architectures and three datasets, and holds zero-shot on UBFC.
Notably, several architectures overshoot the GT reference (VAE-Orth: 0.605;
VAE-Large: 0.535; VQ-VAE: 0.529): the decoder imposes its learned harmonic
prior regardless of input, a signature of template collapse rather than
subject-specific recovery.
The hallucination audit (Section~\ref{sec:experiments}) confirms this is prior
imposition, not rPPG-driven recovery: shuffled rPPG produces indistinguishable
output ($r = {+}0.7178$) from real rPPG ($r = {+}0.7183$), confirming the
H2/H1 improvement does not depend on rPPG input content. The near-constant output H2/H1 across subjects
(standard deviation $\sigma = 0.022$ for VAE-Base vs.\ $\sigma = 0.122$ for GT;
Fig.~\ref{fig:collapse}) confirms that harmonic generation is population-level.

\begin{figure}[t]
\centerline{\includegraphics[width=\columnwidth]{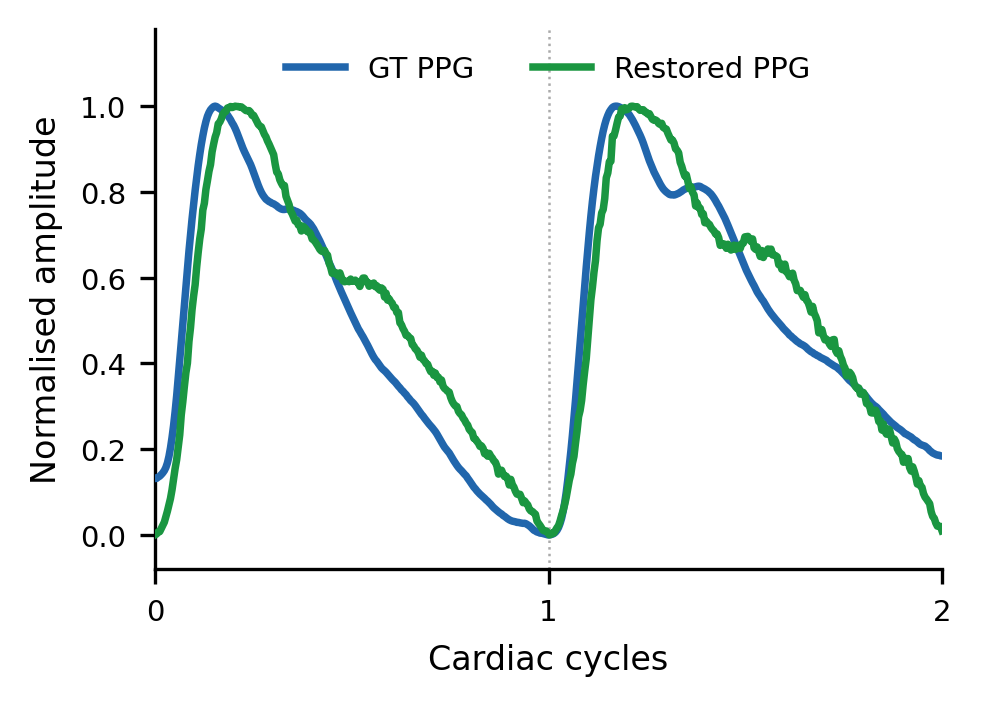}}
\caption{Qualitative illustration of population-level waveform restoration.
Best-accuracy test subject (UBFC-PHYS, VAE-Base per-subject $r = {+}0.870$);
three consecutive highest-quality rPPG cycles (SQI~$= 7.91$).
The restored PPG (green) recovers the systolic peak and diastolic slope structure
of the contact PPG ground truth (blue), while the raw rPPG (red dashed) carries
only noisy cardiac-frequency content. The restored morphology is the
population-level prior, not subject-specific recovery; cross-subject
$r = 0.808$ confirms template collapse persists at the subject level.}
\label{fig:waveformrestoration}
\end{figure}

Fig.~\ref{fig:waveformrestoration} illustrates what this population-level
restoration looks like in practice on the best-accuracy test subject.

The consistency of population-level harmonic generation across all architectures
raises the question of whether the model reads the rPPG input at all, or applies
the prior regardless of input content.

\subsection{Hallucination Gap}

A hallucination audit fed VAE-Orth three input types: real rPPG, white noise,
and temporally shuffled rPPG ($N = 27$ test subjects; per-subject~$r$ under
real rPPG is~0.7183 in this controlled protocol vs.\ 0.681 in the standard
evaluation pipeline, owing to cycle-selection differences between the two
procedures). The mean per-subject~$r$
under real rPPG ($r = {+}0.7183$) exceeded white noise ($r = {+}0.541$) for
18 of 27~subjects; the difference did not reach significance (Mann-Whitney~U,
$p = 0.150$).

\begin{figure}[htbp]
\centerline{\includegraphics[width=\columnwidth]{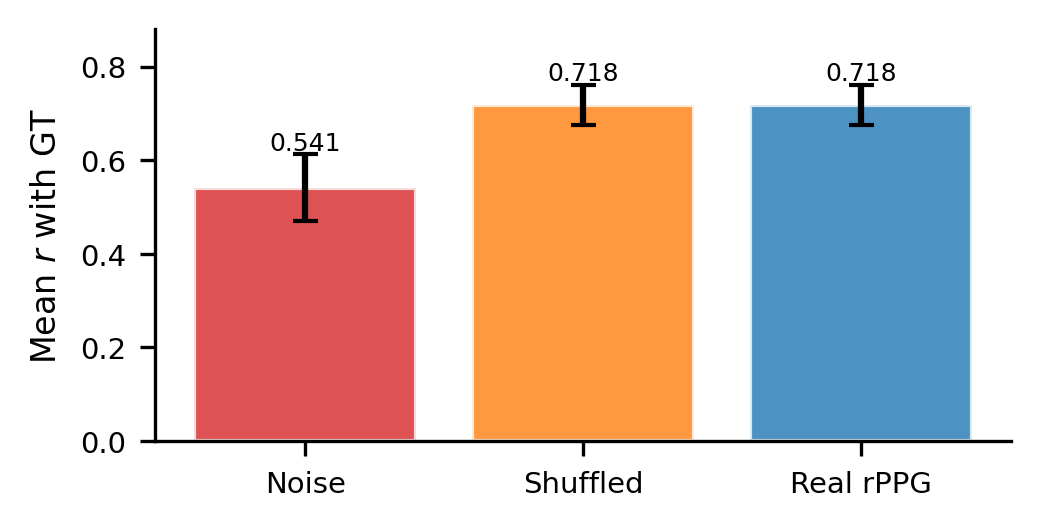}}
\caption{Hallucination gap: mean~$\pm$\,SEM per-subject~$r$ under three input
conditions ($N = 27$ test subjects). Real rPPG ($r = {+}0.7183$) exceeds white noise
($r = {+}0.541$); shuffled rPPG ($r = {+}0.7178$) closely matches real rPPG, indicating
amplitude-distribution dominance. Mann-Whitney~U (MWU) $p = 0.150$ (not significant).}
\label{fig:halluci}
\end{figure}

The final experiment tests whether the morphological prior, trained on
high-frequency contact PPG, transfers to a completely unseen hardware
configuration.

\subsection{Zero-Shot Generalisation}

VAE-Base, trained exclusively on 500\,Hz$+$ Polymate contact PPG, achieves
mean per-subject $r = {+}0.708$ on the 16~UBFC test subjects
(CMS50E 64\,Hz, never seen during training) without any retraining or adaptation.
All 16~subjects show positive per-subject~$r$ ($\sigma = 0.107$; range $0.514$--$0.877$). The successful generalisation from 500\,Hz$+$ Polymate training data to
64\,Hz CMS50E data without retraining demonstrates that the VAE prior
captures hardware-agnostic population-level cardiac structure. The constraint on subject-specific recovery is not sensor
incompatibility; it is the absence of subject-discriminative information in the
rPPG signal that feeds the encoder.

\section{Discussion}
\label{sec:discussion}

Consumer face cameras cannot support subject-specific morphological recovery at
any architecture or loss family: the information is absent from, or present below
the extractable threshold of, the current measurement chain.
This is a physical constraint of the vascular architecture sampled by
green-channel rPPG, not an architectural failure.
Nor is it attributable to an impoverished morphological prior: the VAE was trained on 500--1000\,Hz Polymate contact PPG that fully resolves the dicrotic notch, systolic timing, and harmonic structure, ruling out prior fidelity as an alternative explanation for the null result.
Notably, VAE-Base achieves per-subject $r = {+}0.652$, below the trivial
mean-predictor baseline of~$0.770$: the frozen decoder imposes a morphologically
plausible but canonically smooth waveform approximating the population mean;
because the encoder receives no subject-discriminative signal, this output
cannot match any individual's waveform more closely than the raw mean, and the
smoother canonical shape scores below the raw population mean on per-subject~$r$.
This is corroborating evidence for the null, not a training failure.
Within this limit, three findings characterise what the chain does preserve:
population-level harmonic content, hardware-agnostic generalisation of the
learned prior, and partial signal reading at the amplitude level.
The non-significant hallucination gap ($p = 0.150$) provides directional but
inconclusive evidence of this partial reading; the shuffled rPPG result
($r = {+}0.7178 \approx$ real rPPG $r = {+}0.7183$) is the stronger indicator
that amplitude distribution, not temporal structure, governs model output.
The physical mechanism is detailed below; the ablation confirming that the loss
objective, not the architecture, is the bottleneck is reported in
Section~\ref{sec:experiments}.

The physical explanation lies in the measurement chain. Consumer face cameras
sample only the superficial capillary plexus, whose pulsatile signal is
attenuated and low-pass filtered by the Windkessel mechanism of the
microvasculature~\cite{ref1}. At 30\,fps with the green channel, the dicrotic
notch (50--100\,ms, 1--3~sample points) falls at or below the joint temporal
resolution and SNR floor of the measurement. The arterial-wall displacement
encoding subject-specific stiffness biomarkers is orders of magnitude smaller than
the capillary light absorption change detected by the
camera~\cite{ref1,ref36,ref39}. Motion artifacts further degrade the SNR of the
heart-beat component~\cite{ref37}. The structural divergence between face-camera
rPPG and finger contact PPG waveforms has been directly quantified:
Braun~\textit{et~al.}~\cite{ref55} show that the systolic upstroke slope,
dicrotic notch position and depth, and amplitude distribution differ
substantially between measurement sites, not as a consequence of noise but of
the vascular architecture at each site. The information required for
subject-specific morphological recovery is physically absent from the camera
measurement, regardless of which algorithm is applied.

Our finding is complementary to, and physically distinct from,
Wu~\textit{et~al.}~\cite{ref47}, who achieve contactless diastolic BP estimation
from facial rPPG with RMSE 7.32~mmHg. Their approach extracts inter-ROI
propagation delays and systolic upstroke duration: temporal ratio quantities
computed across spatially separated face regions. These features are timing
relationships, not waveform shapes, and they survive the rPPG measurement chain
because propagation delay is encoded in the relative phase of camera channels
rather than in the absolute morphology of any individual cycle. Our work and
Wu~\textit{et~al.}\ address complementary and physically distinct feature classes.
Sun~\textit{et~al.}~\cite{ref46} warrant a more detailed comparison, as their
biometric authentication result ($\text{EER} = 2.16\%$, 100~subjects) might
appear to contradict our finding. Three distinctions resolve this. First,
Sun~\textit{et~al.}\ authenticate using multi-cycle session averages spanning
5--20 consecutive heartbeats; stable individual characteristics (resting heart
rate, HRV, vasomotor tone, respiratory modulation) can emerge from multi-cycle
averaging and survive the rPPG chain even if single-cycle morphological shape
does not. Second, their hybrid training teaches the rPPG branch a
\textit{consistent} per-subject mapping; consistency is sufficient for
classification but does not require dicrotic notch depth or systolic shape to
be physically recovered from the optical signal. Third, classification requires
only that some consistent signal difference exists between subjects, however
small; reconstruction requires that the full waveform shape be encoded in the
input with sufficient SNR. The rPPG chain may cross the classification boundary
while remaining below the reconstruction boundary. Our result does not
contradict Sun~\textit{et~al.}; it delineates the regime in which their result
cannot be extended.

Standard ML solutions to template collapse presuppose discriminative information
in the input and close shortcuts to using it. BicycleGAN~\cite{ref48} enforces
bijective consistency to prevent a conditional GAN from collapsing all outputs to
one mode, but this works only when the output space is genuinely diverse given the
input. VICReg~\cite{ref49} penalises variance collapse across batches, but the
variance term can only spread representations across structure that is already
present. When $I(\text{rPPG}_{\text{cycle}};\,\text{subject\_morphology}) \approx 0$,
as evidenced by our SupCon null, there is no discriminative structure to spread
across and no shortcut to close. This failure mode is independently corroborated by SIGMA-PPG~\cite{ref45},
which reports that contrastive objectives suppress fine morphological detail in
contact PPG restoration, a finding on wrist contact PPG where the target
information is physically present. On camera rPPG, where the information is
absent, the suppression is total: the contrastive loss converges to the
theoretical null rather than merely degrading fine structure. The two results
together establish a spectrum: contrastive objectives suppress morphological
detail on information-weak inputs and eliminate it entirely on
information-absent inputs.

\begin{table*}[t]
\caption{Taxonomy of ML collapse types and why standard remedies are inapplicable
in the rPPG morphological recovery regime.
For each class, the standard solution presupposes discriminative structure in the
conditioning input. When $I(\text{rPPG}_{\text{cycle}};\,\text{subject\_morphology}) \approx 0$,
as evidenced by SupCon convergence to $\log N = 4.844$ across six variants,
every remedy operates on a null premise.}
\label{tab:taxonomy}
\centering
\renewcommand{\arraystretch}{1.6}
\setlength{\tabcolsep}{4pt}
\begin{tabular}{p{0.12\textwidth}p{0.12\textwidth}p{0.13\textwidth}p{0.16\textwidth}p{0.17\textwidth}p{0.16\textwidth}}
\toprule
\textbf{ML Collapse Type} & \textbf{Mechanism} & \textbf{Standard Solution} &
\textbf{Our Equivalent} & \textbf{Our Result} & \textbf{Why It Fails Here} \\
\midrule
Posterior collapse (VAEs) &
  Decoder ignores latent code $z$ &
  KL annealing, free bits, $\delta$-VAE &
  VAE-Base KL-balanced training &
  VAE prior is sound; collapse is in Stage~2 &
  Input, not architecture \\[3pt]
Mode collapse (GANs) &
  Generator ignores conditioning signal &
  Wasserstein loss, diversity discriminator &
  Two-Stage+Div subject-mean diversity penalty &
  $r_\text{cross}$: $0.960 \to 0.892$ (partial, $-$10.7\% acc.) &
  Hallucinated diversity, not recovered \\[3pt]
Conditional input ignoring (img2img) &
  Output same for all conditioning codes &
  BicycleGAN bijective consistency &
  VAE-Orth orthogonal cascade disentanglement &
  Cross-subj $r$ worsened to 0.970 &
  No bijective structure to exploit \\[3pt]
Representation collapse (contrastive) &
  All representations map to same subspace &
  VICReg variance term, DirectCLR~\cite{ref12} &
  SupCon loss (6~variants, $\lambda = 1$--$20$) &
  All 6~variants: $\log N = 4.844$ (null) &
  $I(\text{rPPG};\,\text{morphology}) \approx 0$; no structure to spread \\[3pt]
Modality collapse (multimodal) &
  Model ignores weaker modality &
  Modality dropout, gradient modulation &
  RGB-FPS camera-only (removes rPPG preproc.) &
  Cross-subj $r = 0.9957$ (worse than hybrid) &
  Weak modality carries no signal to recover \\
\bottomrule
\end{tabular}
\end{table*}

Table~\ref{tab:taxonomy} systematises this finding across the full landscape of
known ML collapse types, mapping each of the 16~architectures to its
corresponding category and solution. KL-balanced VAE-Base addresses posterior
collapse; the Two-Stage+Div diversity penalty is the functional analogue of
Wasserstein-style mode collapse remedies; VAE-Orth orthogonal disentanglement
mirrors BicycleGAN bijective consistency for conditional input ignoring; SupCon
with six variants is the most sensitive possible test for representation
collapse, equivalent to VICReg variance spreading applied to an information
probe; and camera-only encoders (RGB-Window, RGB-Physics, RGB-FPS) represent
the modality dropout approach to modality collapse. None resolved template
collapse. The unified explanation is that every known remedy shares one
implicit assumption: the information to be extracted is present in the
conditioning input and collapse is a training pathology preventing its
extraction. Our SupCon null removes this assumption. When
$I(\text{rPPG}_{\text{cycle}};\,\text{subject\_morphology}) \approx 0$,
collapse is not a training pathology; it is the correct behaviour of a
well-trained model given an uninformative input.

\begin{figure}[htbp]
\centerline{\includegraphics[width=\columnwidth]{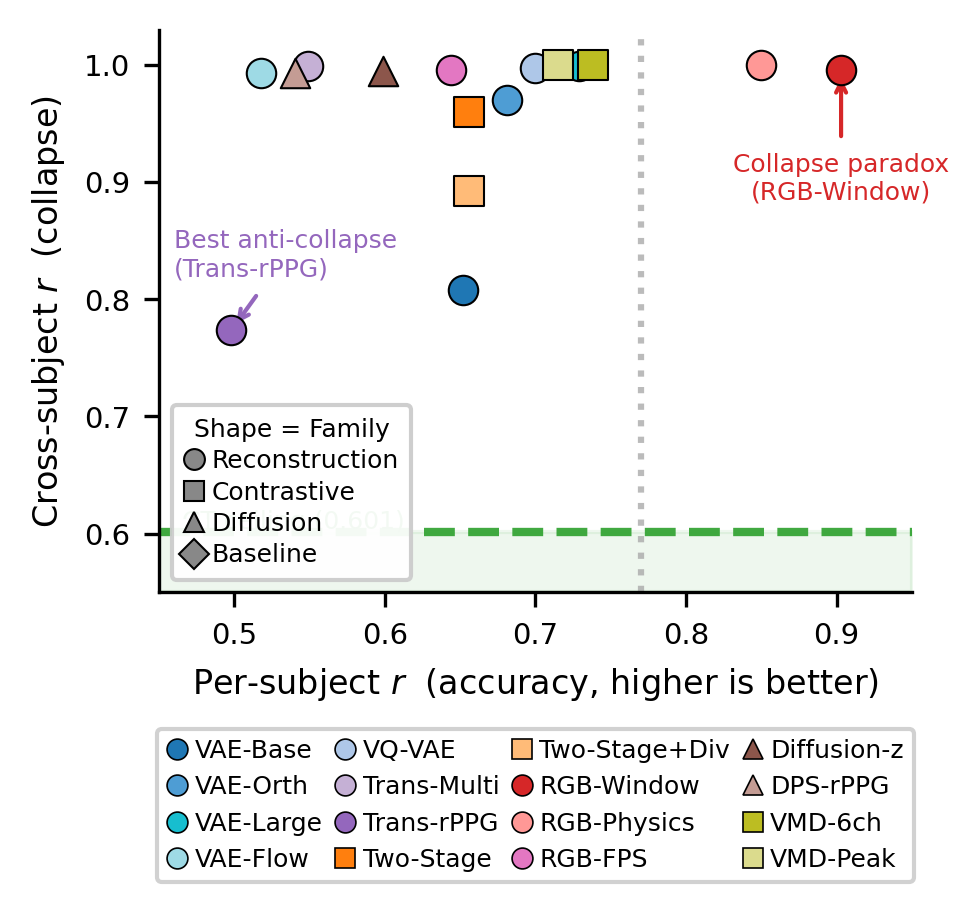}}
\caption{Per-subject~$r$ vs.\ cross-subject~$r$ for all 16~architectures.
Higher per-subject~$r$ (horizontal axis) is better; lower cross-subject~$r$
(vertical axis) indicates less collapse. The green dashed line marks the GT
ceiling ($r_{\text{cross}} = 0.601$); no architecture reaches it.
RGB-Window achieves the highest per-subject~$r = 0.903$ yet sits at
near-total collapse ($r_{\text{cross}} = 0.996$), demonstrating that
per-subject~$r$ alone cannot detect template collapse.}
\label{fig:scatter}
\end{figure}

A critical methodological warning follows from these results. Per-subject
Pearson~$r$ alone is insufficient to evaluate subject-specific waveform
reconstruction. Fig.~\ref{fig:scatter} makes this concrete: RGB-Window achieves
$r = {+}0.903$, the highest value in our study, while simultaneously sitting at
complete template collapse (cross-subject $r = 0.996$; GT ceiling 0.601). A
trivial predictor returning the population mean for every subject achieves
$r = {+}0.770$. Without cross-subject~$r$, prior work reporting high per-subject~$r$
on camera rPPG waveform recovery cannot be distinguished from complete collapse.
We recommend that cross-subject~$r$ be reported alongside per-subject~$r$ as a
standard diagnostic in all future rPPG waveform reconstruction benchmarks.

Three hardware paths offer principled routes beyond the current limit, each
targeting a different physical bottleneck identified by our analysis.
Multi-wavelength and near-infrared acquisition moves beyond the superficial
capillary plexus via two complementary mechanisms: NIR light (${\sim}850$\,nm)
penetrates to subcutaneous depths where arterioles carry the arterial pressure
wave with substantially less Windkessel damping than the capillary
plexus~\cite{ref57}; and Hou~\textit{et~al.}~\cite{ref56} demonstrate that rPPG
signals extracted simultaneously at visible and NIR wavelengths provide
morphological features sufficient for camera-based blood pressure estimation,
establishing the empirical feasibility of this hardware path.
Targeting anatomically accessible superficial vessels changes the measurement
chain at source: Cao~\textit{et~al.}~\cite{ref53} demonstrate that pulsatile
signals from neck blood vessels are measurable contactlessly with a high-speed
camera, confirming that accessible superficial arterial sites carry richer
morphological content than the facial capillary bed sampled by standard rPPG.
Shallow arterial sites such as the temporal artery at 1--2\,mm depth similarly
encode arterial-wall displacement rather than capillary absorption, directly
targeting the vessel-type bottleneck identified in this study.
Multi-site pulse transit time estimation exploits inter-ROI timing delays that do
survive the rPPG measurement chain: Slapni\v{c}ar~\textit{et~al.}~\cite{ref51}
and Shirbani~\textit{et~al.}~\cite{ref52} demonstrate that narrow-band
multi-wavelength camera rPPG and facial video respectively enable remote PTT
estimation, providing a contactless path to arterial stiffness indices without
requiring per-cycle morphological shape recovery.
Algorithm advances alone cannot overcome the fundamental measurement constraint
we have characterised; hardware advances that change the measurement physics are
the necessary path forward.

Three questions remain outside the scope of this study and define the open
frontier. First, we did not directly measure the AC/DC ratio of our rPPG
signals relative to contact PPG; we observe the failure mode without
quantifying the SNR gap that separates the two modalities. Second, and more
fundamentally, it is unknown whether subject-specific morphological information
is \textit{absent} from the optical signal at the skin surface, or
\textit{present but below the extractable threshold} at current SNR; these are
physically distinct situations with different implications for what hardware
improvement would be sufficient. Third, morphological reconstruction from a
contactless camera targeting a known superficial artery (as opposed to diffuse
face illumination) has not been tested. The present study exhausts the
architectural space for green-channel, face-ROI, ambient-light rPPG; it does
not exclude the possibility that a targeted hardware change crosses the
information threshold.
Task-state variation in UBFC-PHYS morphological targets (T1/T2/T3 stress
protocols) is an uncontrolled confound; reconstruction error in UBFC-PHYS
subjects may partly reflect target distribution shift relative to the
resting-state VAE prior.

\FloatBarrier
\section{Conclusion}
\label{sec:conclusion}
SupCon convergence to $\log N = 4.844$ across six independent architectural
variants spanning output space, latent space, and a 20$\times$ temperature
range constitutes the strongest available empirical evidence that no discriminative
morphological structure is extractable from single-cycle rPPG by the encoder
families tested: a physical limit of the measurement chain, not an
architectural failure.
Cross-subject~$r$, introduced here as a collapse diagnostic, reveals complete
template collapse in architectures that appear accurate by per-subject~$r$ alone;
without it, the field cannot distinguish genuine subject-specific recovery from
population-mean prediction.
Population-level harmonic content is recoverable from consumer camera data and
generalises zero-shot across sensor hardware; the VAE prior is sound and
hardware-agnostic, and the binding constraint is the measurement chain, not the
model.
Standard ML anti-collapse solutions presuppose discriminative structure in the
conditioning input; when that structure is absent, they produce hallucinated
diversity rather than recovered morphology, a finding applicable to any sensing
task where the measurement chain eliminates the target signal.
The cross-subject~$r$ metric and physical limits established here provide the
reference benchmark against which future hardware advances in non-invasive
cardiovascular monitoring can be evaluated.

\appendices



\end{document}